\documentclass{ieeeaccess}
\usepackage{cite}
\usepackage{amsmath,amssymb,amsfonts}
\usepackage{graphicx}
\usepackage{textcomp}

\usepackage{algorithm}
\usepackage{listings}
\usepackage{lipsum}
\usepackage{multirow}
\usepackage{lineno,hyperref}
\usepackage{algpseudocode}
\usepackage{booktabs}
  
\usepackage{caption}
\DeclareCaptionFont{ieeeblue}{\color{accessblue}}
\DeclareCaptionLabelFormat{myformat}{\raggedright\figcapfont{\textbf{#1}\textbf{#2}}}
\def \hcaption {7pt}

\def\BibTeX{{\rm B\kern-.05em{\sc i\kern-.025em b}\kern-.08em
    T\kern-.1667em\lower.7ex\hbox{E}\kern-.125emX}}
\begin{document}

\history{Received July 15, 2019, accepted August 16, 2019, date of publication August 26, 2019, date of current version September 12, 2019.}
\doi{ 10.1109/ACCESS.2019.2937505}

\title{A System for Automatic English Text Expansion}
\author{\uppercase{Silvia Garc\'ia-M\'endez}\authorrefmark{1},
\uppercase{Milagros Fern\'andez-Gavilanes\authorrefmark{2}, Enrique Costa-Montenegro\authorrefmark{1}, Jonathan Juncal-Mart\'inez\authorrefmark{1}, Francisco J. Gonz\'alez-Casta\~no\authorrefmark{1}, and Ehud Reiter\authorrefmark{3}}}
\address[1]{GTI Research Group, Telematics Engineering Department, School of Telecommunications Engineering, University of Vigo, 36310 Vigo, Spain}
\address[2]{Defense University Center, 36920 Mar\'in, Pontevedra, Spain}
\address[3]{CLAN Research Group, Department of Computing Science, Meston Building, University of Aberdeen, Aberdeen AB24 3UE, Britain}
\tfootnote{This work was partially supported by grants TEC2016-76465-C2-2-R (Mineco, Spain), GRC-2018/53 and ED341D R2016/012 (Xunta de Galicia, Spain), and a University of Vigo travel grant to visit the CLAN Research Group at the University of Aberdeen in Scotland.}

\markboth
{Silvia Garc\'ia-M\'endez \headeretal: A System for Automatic English Text Expansion}
{Silvia Garc\'ia-M\'endez \headeretal: A System for Automatic English Text Expansion}

\corresp{Corresponding author: Silvia Garc\'ia-M\'endez (e-mail: sgarcia@gti.uvigo.es).}

\begin{abstract}
We present an automatic text expansion system to generate English sentences, which performs automatic Natural Language Generation ({\sc nlg}) by combining linguistic rules with statistical approaches. Here, ``automatic" means that the system can generate coherent and correct sentences from a minimum set of words. From its inception, the design is modular and adaptable to other languages. This adaptability is one of its greatest advantages. For English, we have created the highly precise aLexiE lexicon with wide coverage, which represents a contribution on its own. We have evaluated the resulting {\sc nlg} library in an Augmentative and Alternative Communication ({\sc aac}) proof of concept, both directly (by regenerating corpus sentences) and manually (from annotations) using a popular corpus in the {\sc nlg} field. We performed a second analysis by comparing the quality of text expansion in English to Spanish, using an ad-hoc Spanish-English parallel corpus. The system might also be applied to other domains such as report and news generation.
\end{abstract}


\titlepgskip=-15pt

\maketitle

\section{Introduction}
\label{intro}

Natural language generation ({\sc nlg}) systems \cite{ReiterDale00} take quantitative, visual and linguistic data (words, sentences, texts) as input. In particular, we are interested in text-to-text expansion, which generates complete sentences or even texts from some meaningful words. A text expansion system must add elements like conjunctions and prepositions to transform the input words into linguistically correct outputs. For example, given the input words `she', `look', `picture', `yesterday', `not', the output might be `She did not look at the picture yesterday'. This is useful in Augmentative and Alternative Communication ({\sc aac}) \cite{elsahar2019augmentative}, for instance. In this case, a user would select pictograms corresponding to input words on a tablet to obtain the equivalent linguistically correct written and oral output.

Our approach, which is modular and extensible to other languages, is based on our previous work on automatic {\sc nlg} systems for Spanish \cite{silvia2018web4all,garcia2019library}. We have improved modularity that allows to easily replace domain-dependent components.

The rest of this paper is organised as follows. In Section \ref{related_work} we review the state-of-the-art, paying special attention to English {\sc nlg}, automatic generation and text expansion. In Section \ref{methodology} we describe our contribution, a system for English text expansion based on the aLexiE English lexicon (Section \ref{lexicon}), a grammar (Section \ref{grammar}), and a sentence planner and a surface realiser (Section \ref{nlg_system}). In Section \ref{evaluation} we present our evaluation results, based on a widely used corpus in the {\sc nlg} field (Section \ref{aac-corpus}) and an ad-hoc Spanish-English parallel corpus (Section \ref{paralel-corpus}). Finally, Section \ref{conclusions} concludes the paper.

\section{Related work}
\label{related_work}

{\sc nlg} tasks generally follow well-defined sub-tasks \cite{ReiterDale00, ReiterDale1997}: content determination, text structuring, sentence aggregation, lexicalisation (expressing information with the right words), referring expression generation (domain objects identification), and linguistic realisation (text correctness). 

Content determination filters irrelevant information. It is obviously context- and application-dependent. For this sub-task, it has been proposed to apply data-driven techniques \cite{kutlak13}. 

Text structuring plans the discourse and sets the sentence presentation order. Recent solutions are based on manual rules \cite{mairesse2007personage,duvsek2015training} and Systemic Functional Grammar ({\sc sfg}) \cite{Bateman97}, Rhetorical Structure Theory ({\sc rst}) \cite{williams2008generating}, Meaning-Text Theory ({\sc mtt}) \cite{wanner2010marquis}, Machine Learning \cite{lampouras2016imitation,mei2016talk}, and the Centred Theory \cite{barzilay2008modeling}, to cite some. 

Semantic- and syntactic-level sentence aggregation join data into single fluent and readable sentences \cite{walker2007individual}.

Lexicalisation transforms the result of sentence aggregation into {\it natural language} ({\sc nl}) but it must choose between different expression alternatives in {\sc nl}. Generally, results improve with more alternatives \cite{hervas2005case}.

Referring Expression Generation ({\sc reg}) selects the phrases or words that can unambiguously describe domain entities. It selects the best properties to distinguish elements, and it discards irrelevant information. {\sc reg} algorithms include the Full Brevity Algorithm \cite{dale1989cooking}, the Greedy Heuristic \cite{dale1992generating,frank2009using}, and the Incremental Algorithm \cite{dale1995computational}. 

Linguistic realisation comprises generation of morphological forms as well as insertion of auxiliary verbs, prepositions and punctuation marks. It may fall into the {\it generation gap}, because input data are often incomplete (i.e. they lack elements to be syntactically comprehensible) \cite{meteer1991bridging}. Templates avoid this gap and ensure consistency. However, using templates to automatically transform data into text (which are typically used in applications such as weather, traffic, sports, and health reporting) only yields better results than other approaches in small domains with little variation. Hand-coded grammar-based systems outperform templates when detailed input is available as in the case of {\sc kpml}\footnote{Available at {\tt http://www.fb10.uni-bremen.de/anglistik /langpro/kpml/README.html}, July 2019.} \cite{Bateman97}. There are other alternatives like statistical methods that produce probabilistic grammars from large corpora increasing coverage at less effort \cite{Langkilde02}; such as the Head-driven Phrase Structure Grammar ({\sc hpsg}) \cite{nakanishi2005probabilistic}; the Lexical-Functional Grammar ({\sc lfg}) \cite{cahill2006robust}; the Tree-Adjoining Grammar ({\sc tag}) \cite{gardent2015multiple}, and some Deep Learning approaches \cite{wen2015semantically}. We employ a hybrid system that combines the advantages of stochastic and grammar-based systems with low {\sc nlg} complexity since it uses only keywords to generate complete sentences.

Intelligent {\sc nlg} architectures may be modular, (roughly following the previous stages of macro-planning and text structuring, micro-planning or sentence aggregation, lexicalisation, referring expression generation, and linguistic realisation with syntactic and morphological rules); planning-oriented (which are less modular); or data-driven, based on statistical learning. Rule- (or template-) based approaches \cite{cheyer2007method,mirkovic08}, however, are the most extended nowadays in real applications. 

According to types of output texts, these may be informative texts, summaries, dialogues, recommendations, and persuasions or creative writings. Most informative text systems generate routinary information from quantitative data \cite{Reiter95,dale2005using,liu2017automatic}. Summary generation \cite{mani2001automatic} has applications in areas such as medicine, sports, and finance. Persuasive texts are intended to shape user behaviour \cite{Reiter03}. Dialogue systems, of interest for call centres or gaming interfaces, focus on human-machine communication \cite{Fiedler05,morris2002conversational}. Creative text generation is extremely difficult. It has been demonstrated that predefined templates are too rigid for it \cite{Peinado04}. Affective {\sc nlg} systems have been able to generate texts beyond factual information such as poetry \cite{Gervas2001}, but the operator has no control over the process. The quality of the results in {\sc nlg} is measured in terms of adequacy, fluency, readability, and variation \cite{Stent05}. There is a trade-off between efficiency and output quality.

Next we review the most relevant existing systems. SimpleNLG \cite{Gatt09} performs surface realisation by following a knowledge-based approach. Originally for English, it is now available for German \cite{bollmann2011adapting}, French \cite{VaudryLapalme13}, Brazilian Portuguese \cite{de2014adapting}, Italian \cite{mazzei2016simplenlg}, Spanish \cite{garcia2017natural, soto2017adapting}, Dutch \cite{simplenlgdutch}, Mandarin \cite{simplenlgmandarin}, and Galician \cite{fuentes2018adapting}. This library has had strong impact in {\sc nlg} due to its simple usage. Its main disadvantage from the perspective of our research is its manual operation. The NaturalOWL \cite{Androutsopoulos14} data-to-text manual tool imposes a complex input format. It generates texts from the {\sc owl} ontology. SumTime \cite{Reiter05,Belz08}, also for data-to-text generation, is highly sensitive to language variations
and, thus, it is only adequate for the language it was designed for. 

The system in \cite{chen2002towards} comprises a trainable sentence planner and a probabilistic surface realiser. Its modular design is similar to ours, but it is only available for English and it is not easily adaptable to other languages since certain language-dependent resources are required to train the surface realiser (a {\sc tag} grammar, the source of the supertags to annotate the semi-specified {\sc tag} derivation tree, a treebank to obtain the tree stochastic model driving the tree chooser and a corpus of sentences to train the language model required for the linear precedence chooser). Our system also needs two language-dependent elements, the lexicon and the grammar, but they can be easily replaced and we provide enough information to create/adapt these elements to any domain.

In the particular field of text expansion in English applied for {\sc aac}, we must mention the work by \cite{Demasco1992}, in line with automatic generation. Their {\it sentence compansion} technique \cite{demasco1989towards,mccoy1990applying}\footnote{We thank Kathleen F. McCoy for informing us about project {\sc compansion.}} takes a compressed message and expands it into a well-formed sentence. In practice it is useful as a writing assistant or a conversational aide in situations where grammatically correct output is desired. Its bottleneck is its generation method, based on a unification-type grammar that needs to explore many possibilities to deliver its output. Thus, it is highly time consuming. Besides, most of its operation is based on markers linked to a lexicon, as in the case of plural forms and possessives (the system might interpret the noun `apple' followed by a plural marker as `apples'; the pronoun `I' followed by a possessive marker would be interpreted as `my'). Input words must be present in the lexicon, since they are taken from a word board and the user is not free to enter them. To avoid this constraint we have designed an automatic procedure for lexicon acquisition. It does not need markers, so that the entire generation process is much faster.

Summing up, most {\sc nlg} systems are purpose-built and, as such, they are highly sensitive to problem characterization. On the contrary, thanks to the modularity of our system we can isolate domain-dependent modules (grammar and lexica) from domain-independent ones ({\sc nlg} surface realiser). It can be tailored to different applications and domains using the corresponding syntactic structures and vocabularies. It can be easily extended to other languages as well.

As far as we know there are no other systems for automatic text-to-text expansion in English with a hybrid architecture.

\section{Methodology and architecture of the proposed system}
\label{methodology}

\subsection{aLexiE lexicon}
\label{lexicon}

This section describes the morphological part, that is, the lexicon providing linguistic knowledge. This is our first contribution.

We pursue a fully automatic {\sc nlg} system. It must select the grammar structure for the input words and their inflection. Therefore we need an ample vocabulary with linguistic data. The aLexiE lexicon serves this purpose. We created it by interpreting input resources and automatically (without human supervision) merging them with the two-step methodology in \cite{CrouchKing05} followed by a final verification step (similarly to \cite{garcia2019library}):

\begin{enumerate}
\item Extraction of all possible entries and translation to a common format (Algorithm \ref{alg:stage1}).
\item Automatic comparison and combination of existing lexica to create the new resource (graph unification in \cite{necsulescu2011towards} and \cite{BelEtAl11}).
\item Lexical verification of extracted and translated entries and their categories against the Merriam-Webster Dictionary\footnote{Available at {\tt https://www.merriam-webster.com}, July 2019.} ({\sc mwd}) (Algorithm \ref{alg:stage2}). 
\end{enumerate}

\subsubsection{Linguistic resources and creation of aLexiE}
\label{sub-linguistic-resources-procedure}

We built aLexiE from free English linguistic resources. We prioritized correctness and coverage and selected the following:

\begin{itemize}
\item The morphological and syntactic English lexicon from the Alexina Project\footnote{Available at {\tt https://gforge.inria.fr/projects/alexina}, July 2019.} (EnLex) \cite{Sagot10}.

\item The Specialist Lexicon\footnote{Available at {\tt https://lsg3.nlm.nih.gov/LexSysGroup/Pro-
jects/lexicon/current/web/index.html}, July 2019.} ({\sc nih}) of medical terms and everyday words.

\item The Freeling English dictionary\footnote{Available at {\tt http://nlp.lsi.upc.edu/freeling/index.php/
node/12}, July 2019.} ({\sc en-freeling}), automatically extracted from {\sc wsj} (Wall Street Journal) and other corpora.
\end{itemize}
 
We performed extraction and mapping independently. Once the information was taken from the selected resources, it was transformed to a common format. Unlike for the Spanish version in \cite{garcia2019library}, input resources were unrelated, and thus we conducted independent extraction and translation stages for each selected resource.

We first extracted entries from EnLex tagged as noun, pronoun, verb, adjective, adverb, determiner, conjunction, and preposition, ignoring interjections, numerals, and proper nouns. Each resulting EnLex word entry was translated to the extensional Alexina format \cite{DanlosSagot08}. As an example, Listing \ref{picture-resources} illustrates the EnLex entry for the English lemma `picture'. It can be observed that this noun (cat=n), has two forms, `picture' and `pictures', respectively for masculine singular and plural.

\begingroup
\newlength{\xfigwd}
\setlength{\xfigwd}{\textwidth}
\begin{lstlisting}[caption={\vspace{\hcaption}Example of the English lemma `picture' in EnLex.}, label={picture-resources}]
picture	N2 100;Lemma;N;;cat=N;
%default #dela+multext+init
<table name="N2" rads=".*">
 <form suffix=" tag="s"/>
 <form suffix="s" tag="p"/>
</table>
\end{lstlisting}
\endgroup

We did the same for the other two linguistic resources that we selected: {\sc nih}\footnote{We extracted adjectives, adverbs, conjunctions, determiners, prepositions, and pronouns from this resource. There was not inflection information in the entries.} and the Freeling English Dictionary\footnote{From this resource we extracted adjectives, adverbs, and verbs.}.

We kept the present, past, present participle, and past participle forms of English verbs. This information allowed adjusting the verbal tense to context-dependent semantic features. In the case of adjectives, we did not save the comparative and superlative forms (we leave comparative and superlative clauses to future work).

The merging process requires to handle the issue of the different formats and tags of word entries in the selected resources. Algorithm \ref{alg:stage1} converts them to a common format (note: $e$ is an entry in a lexicon).

\begin{algorithm}[H]
\small
 \caption{\label{alg:stage1}: {\bf Extraction and conversion algorithm}}
 \begin{algorithmic}[0]
 \scriptsize
 \Function{extraction\_mapping}{\{$\mbox{\sc EnLex}$\},\{$\mbox{\sc nih}$\},\{$\mbox{\sc en-freeling}$\}}
 \For{$e_{\mbox{\scriptsize \sc EnLex}} \in \{\mbox{\sc EnLex}\}$}
 \State lemma$_{e_{\mbox{\scriptsize \sc EnLex}}}$ = $e_{\mbox{\scriptsize \sc EnLex}}$.getLemma()
 \State category$_{e_{\mbox{\scriptsize \sc EnLex}}}$ = $e_{\mbox{\scriptsize \sc EnLex}}$.getCategory()
 \EndFor
 \For{$e_{\mbox{\scriptsize \sc nih}} \in \{\mbox{\sc nih}\}$} \State lemma$_{e_{\mbox{\scriptsize \sc nih}}}$ = $e_{\mbox{\scriptsize \sc nih}}$.getLemma()
 \State category$_{e_{\mbox{\scriptsize \sc nih}}}$ = $e_{\mbox{\scriptsize \sc nih}}$.getCategory()
 \EndFor 
 \For{$e_{\mbox{\scriptsize \sc en-freeling}} \in \{\mbox{\sc en-freeling}\}$} \State lemma$_{e_{\mbox{\scriptsize \sc en-freeling}}}$ = $e_{\mbox{\scriptsize \sc en-freeling}}$.getLemma()
 \State category$_{e_{\mbox{\scriptsize \sc en-freeling}}}$ = $e_{\mbox{\scriptsize \sc en-freeling}}$.getCategory()
 \EndFor 
 \State return({\{$\mbox{\sc EnLex}$\},\{$\mbox{\sc nih}$\},\{$\mbox{\sc en-freeling}$\}})
 \EndFunction
 \end{algorithmic}
\end{algorithm}

Verification in Algorithm \ref{alg:stage2} checks the quality of the word entries. It looks for each lemma and its lexical categories in {\sc mwd} (this dictionary has the advantage that it allows more web queries than other freely available online dictionaries).

Finally, collected entries are merged in a combination step that applies the graph unification in \cite{necsulescu2011towards} and \cite{BelEtAl11}. This operation validates common information by integrating data of different nature and discarding inconsistent information. Specifically,

\begin{enumerate}
\item It joins all entries with a common lemma (homography is only considered for different lexical categories).

\begin{enumerate}
\item For the entries that results from (1), feature structures are unified.
\item Next, a new aLexiE entry is created with these structures. The entry comprises all common information plus any particular data in the source entries.
\end{enumerate}

\item A new aLexiE entry is created for any lexical entry that cannot be generated by combining entries from other lexica.
\end{enumerate}

\begin{algorithm}[H]
\small
 \caption{\label{alg:stage2}: {\bf Verification algorithm}}
 \begin{algorithmic}[0]
 \scriptsize
\vspace{-0.05cm}
 \Function{verification}{\{$\mbox{\sc set}$\}}
 \For{$e_{\mbox{\scriptsize \sc set}} \in \{\mbox{\sc set}\}$}
 \State lemma$_{e_{\mbox{\scriptsize \sc set}}}$ = $e_{\mbox{\scriptsize \sc set}}$.getLemma()
 \State category$_{e_{\mbox{\scriptsize \sc set}}}$ = $e_{\mbox{\scriptsize \sc set}}$.getCategory()
 \If{(lemma$_{e_{\mbox{\scriptsize \sc set}}}$.isNOTin{\mbox{\sc mwd}}()$\ \mbox{\sc or}\
$category$_{e_{\mbox{\scriptsize \sc set}}}$.isNOTin{\mbox{\sc mwd}())
 }\\\hspace{0.7cm}}
 \State \{$\mbox{\sc set}$\}.delete(e$_{\mbox{\scriptsize \sc set}})$
 \EndIf
 \EndFor
 \State return({\{$\mbox{\sc set}$\}})
 \EndFunction
 \end{algorithmic}
\end{algorithm}

We remark that the common extraction and translation format avoids inconsistencies in this merging procedure. Algorithm \ref{alg:building} combines all steps in this section so far.

\begin{algorithm}[H]
\scriptsize
 \caption{\label{alg:building}: {\bf Lexicon building algorithm}}
 \begin{algorithmic}[0]
 \State \{$\mbox{\sc EnLex}$\}= LoadEnLex()
 \State \{$\mbox{\sc nih}$\}= LoadNIH()
 \State \{$\mbox{\sc en-freeling}$\}= LoadEn-Freeling()
 \State $\mbox{{\sc extraction\_mapping}(\{{\sc EnLex}\},\{{\sc nih}\},\{{\sc en-freeling}\})}$
 \State \{$\mbox{\sc aLexiE}$\} = \{$\mbox{\sc EnLex}$\} $\cup$ \{$\mbox{\sc nih}$\} $\cup$ \{$\mbox{\sc en-freeling}$\}
 \State $\mbox{{\sc verification}(\{aLexiE\})}$
 \end{algorithmic}
\end{algorithm}

Therefore, aLexiE was built from inputs extracted from previously existing resources, which were merged into a common format and finally verified. Listing \ref{picture-noun} shows the result for the lemma `picture'.

Note that, in Listing \ref{picture-noun}, this lemma is semantically tagged as an object. We used the Multilingual Central Repository\footnote{A lexical database that integrates the Spanish WordNet into the EuroWordNet framework, see {\tt http://adimen.si.ehu.es/web
/MCR}, July 2019.} ({\sc mcr}) \cite{GonzalezEtAl12} to get the semantic classification of nouns in aLexiE. Algorithm \ref{alg:mcr} summarises this procedure.

Due to its size, indexing aLexiE allows our system to conduct the whole {\sc nlg} process much more quickly.

\subsubsection{Automatic extension of the lexicon}
\label{sec:extension-lexicon}

\begin{figure*}[!ht]
\centering
\includegraphics[scale=0.2]{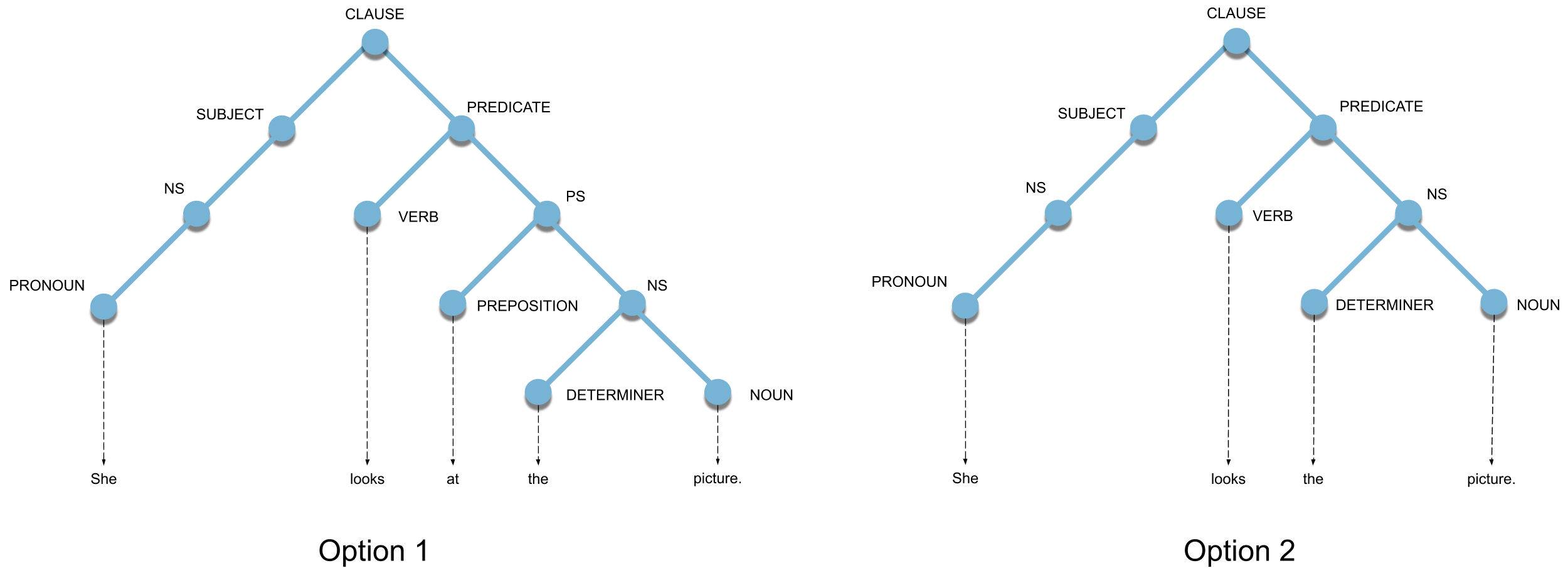}
\caption{\label{syntaxTree} Syntax tree resulting from the grammar}
\end{figure*}

{\sc nlg} can be simplified by avoiding inputs with little meaning, such as prepositions. Consequently, it is 
necessary to infer a priori which preposition follows a particular verb. Indeed, a major challenge in text expansion is inferring missing prepositions. We trained this process from the text in the English Wikipedia, which was previously {\sc pos}-tagged with Spacy Tagger\footnote{Available at {\tt https://spacy.io}, July 2019.}. The language model for this training was based on trigrams centered around verbs, using syntactic and semantic information.

\begingroup
\setlength{\xfigwd}{\textwidth}
\begin{lstlisting}[caption={\vspace{\hcaption}Example of the English lemma `picture' in aLexiE.}, label={picture-noun}]
<?xml version="1.0" encoding="UTF-8"
standalone="no"?>
<lexicon>
 <word>
  <lemma>picture</lemma>
  <category>noun</category>
  <number>singular</number>
  <plural>pictures</plural>
  <semantic_tag>object</semantic_tag>
 </word>
</lexicon>
\end{lstlisting}
\endgroup

\begin{algorithm}[H]
\small
 \caption{\label{alg:mcr}: {\bf Adding syntactic and semantic data}}
 \begin{algorithmic}[0]
 \scriptsize
\vspace{-0.05cm}
 \Function{Add\_Syntatic\&Semantic\_Data}{\{$\mbox{\sc alexie}$\}}
 \For{$e_{\mbox{\scriptsize \sc alexie}} \in \{\mbox{\sc alexie}\}$}
 \State lemma$_{e_{\mbox{\scriptsize \sc alexie}}}$ = $e_{\mbox{\scriptsize \sc alexie}}$.getLemma()
 \State category$_{e_{\mbox{\scriptsize \sc alexie}}}$ = $e_{\mbox{\scriptsize \sc alexie}}$.getCategory()
 \If{(category$_{e_{\mbox{\scriptsize \sc alexie}}}$.isNoun()$\ \mbox{\sc and}\
$lemma$_{e_{\mbox{\scriptsize \sc alexie}}}$.isIn{\mbox{\sc mcr}())
 }\\\hspace{0.7cm}}
 \State $e_{\mbox{\scriptsize \sc mcr}}$ = searchInMCR(lemma$_{e_{\mbox{\scriptsize \sc alexie}}}$)
 \State \{$\mbox{\sc mcr}$\}.add($e_{\mbox{\scriptsize \sc mcr}}$)
 \EndIf
 \EndFor
 \EndFunction
 \end{algorithmic}
\end{algorithm}

As previously mentioned, we used {\sc mcr} to get the semantic classification of the entries tagged as nouns in aLexiE. In this case we established four semantic categories to start working with: living things, foodstuff, places, and objects. For each verb lemma in the training set we analysed if it was followed by a preposition and a noun or a determiner and a noun. We computed the probability by semantic category. Let us consider the entries in Table \ref{languagemodel}. Regarding the verb lemma `look' and the semantic tag {\tt object}, the preposition `at' has the highest probability to go in between according to Table \ref{languagemodelprop}.

\begin{table*}[!ht]
\centering
\caption{\label{languagemodel} Automatic lexicon extension. Sentence examples}
\begin{tabular}{ll}
\hline
\bf Sentence & \bf After processing\\ \toprule
I look the business. & {\tt look + EMPTY + object}\\\hline
She looks at the picture. & {\tt look + at + object}\\\hline
You look at the car. & {\tt look + at + object}\\\hline
She looks like her mum. & {\tt look + like + living thing}\\\bottomrule
\end{tabular}
\end{table*}

\begin{table}[!ht]
\centering
\caption{\label{languagemodelprop} Automatic lexicon extension probabilities by lemma verb and semantic category}
\begin{tabular}{ll}
\hline
\bf After processing & \bf Probability\\ \toprule
{\tt look + EMPTY + object} & 0.33\\\hline
{\tt look + at + object} & 0.66\\\hline
{\tt look + like + living thing} & 1\\\bottomrule
\end{tabular}
\end{table}

In this way, the language model along with the semantic classification allow us to infer the most suitable preposition after a verb by applying semantic knowledge rather than by only considering morphological forms.

Listing \ref{look-verb} shows the aLexiE entry for verb `look'. Note how the language model has learned to add the preposition `at' when the verb is followed by an object (semantically speaking) or `for' in the case of foodstuff or a place. If `look' is followed by a living thing, the system will add the preposition `like'. In the running example in this section, since `picture' is tagged as an object in aLexiE, and given the syntactic and semantic data in the `look' entry; the system will insert preposition `at' between the two words if provided as input to the system.

\subsection{Syntactic structure supported by a grammar}
\label{grammar}

In this section we describe the syntactic stage of our system. It performs syntactic structuring with the Definite-Clause Grammar ({\sc dcg}) \cite{maggiori13}

Syntactic structuring, also called parsing, creates the tree structure of the desired target sentence. We infer this structure by checking the syntactic trees from the grammar for the input words. Obviously diverse possible syntactic structures may result, depending on the roles of the input words in the sentence. 
Fig. \ref{syntaxTree} shows the syntax trees from the grammar for the input words `she', `look', `picture'; `She looks at the picture' and `She looks the picture'. Context-Sensitive Languages ({\sc csl}) are created from this type of grammar. The system picks the most appropriate trees for the input words given the different possibilities within a grammar. 

For the English case, we adapted the simple grammar with wide range of basic sentences described in \cite{garcia2019library}, by adding all its linguistics features, such as adjectives preceding nouns. The system can parse sentences regardless of their complexity. Sentence types may be affirmative, negative, interrogative (in positive or negative form) or imperative (in positive or negative form), including some of the following features: a nominal syntagm subject; a coordinated nominal syntagm subject (compound subject); a nominal syntagm direct complement; a coordinated nominal syntagm direct complement; an indirect complement, and other place or time complements.

\begingroup
\setlength{\xfigwd}{\textwidth}
\begin{lstlisting}[caption={\vspace{\hcaption}Example of the English lemma `look' in aLexiE.}, label={look-verb}]
<?xml version="1.0" encoding="UTF-8"
standalone="no"?>
<lexicon>
 <word>
  <lemma>look</lemma>
  <category>verb</category>
  <present3s>looks</present3s>
  <past>looked</past>
  <present_participle>
   looking
  </present_participle>
  <past_participle>
   looked
  </past_participle>
  <object>at</object>
  <foodstuff>for</foodstuff>
  <living>like</living>
  <place>for</place>
 </word>
</lexicon>
\end{lstlisting}
\endgroup

\begin{figure*}[ht!]
 \centering
 \includegraphics[scale=0.25]{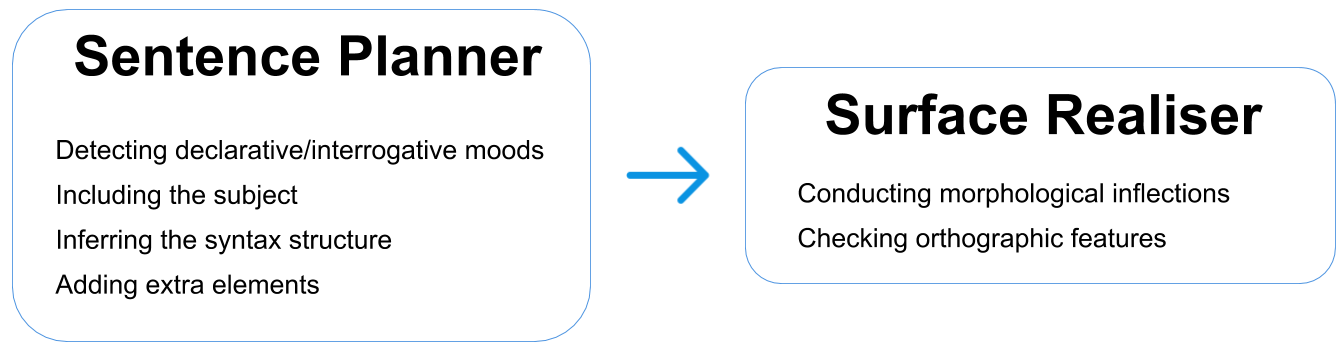}
 \caption{Our two-stage NLG architecture}
 \label{fig:nlg}
\end{figure*}

In our notation, upper case corresponds to tree structures and lower case to word components. Fig. \ref{syntaxTree} illustrates some linguistic rules taken from the grammar.

\subsubsection{Rules for nominal/coordinated nominal syntagms}

Nominal syntagms ({\sc ns}) contain nouns, pronouns or proper nouns. Determiners may precede nouns. A coordinated syntagm is a sentence with two nominal syntagms connected by a conjunction.

\subsubsection{Rules for adjectival/adverbial/prepositional syntagms}

In this case, an adjectival or adverbial syntagm, which consists of an adverb followed by an adjective or vice versa, may precede or follow a noun. Noun-Noun modifiers such as in `car door' are not considered. In prepositional syntagms ({\sc ps}), unless empty, a preposition precedes a nominal syntagm. Prepositional syntagms just follow (never precede) a nominal syntagm.

\subsubsection{Predicate rule}

The sentence predicate contains a verb that may be followed by a nominal syntagm (coordinated or not). The verb may be accompanied by an adjectival/adverbial syntagm. Note that a verb can be followed by two nominal syntagms (yet not coordinated ones, that is, without a conjunction in between), as in sentence `She gave me a cookie yesterday'.

\subsubsection{Sentence rule}

Sentences are composed either of a nominal or coordinated nominal syntagm (subject) and a predicate, or of a single predicate (without subject). The latter is quite common in imperative English clauses. Given the relations among syntagms, the depth level of our system is limited to two iterations to reduce computational load. For example, in case a nominal syntagm includes a prepositional syntagm, the second nominal syntagm cannot contain another prepositional syntagm (to avoid recursion).

\subsection{Proposed NLG library: Sentence Planner and Surface Realiser}
\label{nlg_system}

The input words for our {\sc nlg} library should be meaningful, such as adjectives, nouns, and verbs. The library can automatically infer the determiners, conjunctions, and prepositions that complement those input words in the output sentence.

\begin{figure*}[ht]
\centering
\includegraphics[scale=0.19]{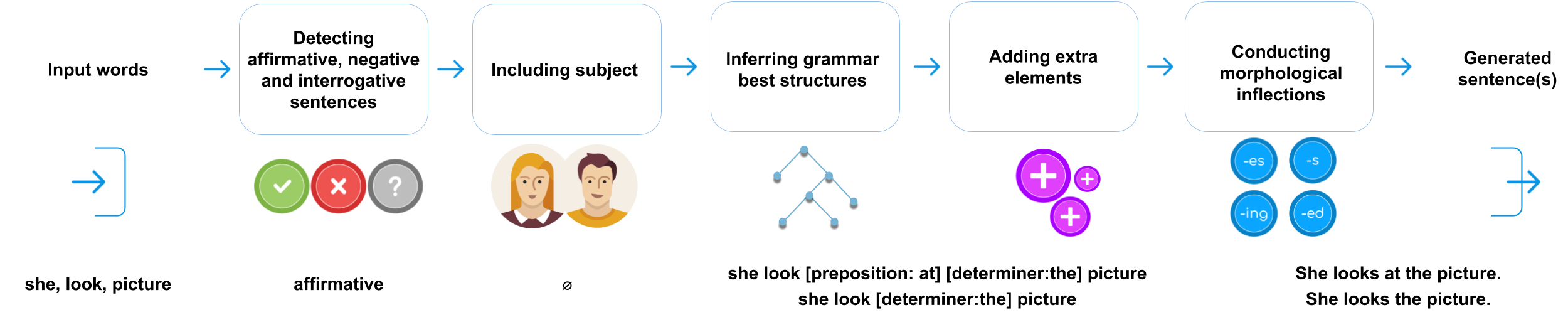}
\caption{\label{flowchart} Flow chart of the NLG procedure}
\end{figure*}

Fig. \ref{fig:nlg} shows our two-stage architecture, an automatic {\sc nlg} processing pipeline. The user introduces the words (plus symbol ? for interrogative sentences) in [subject, verb, object] ({\sc svo}) order, which is not limiting in practice in many domains. The first Sentence Planner stage performs lexicalisation, which adds words and configures sentences. The second Surface Realiser stage, our main contribution, introduces any extra elements that may be necessary and applies morphology inflections to produce grammatically correct and coherent sentences. Fig. \ref{generation_diagram} represents an example of generation using the library. Fig. \ref{flowchart} shows the flowchart of our {\sc nlg} library. The main tasks are the following ones:

\begin{figure*}[ht!]
\centering
\hspace{0.5cm}
\includegraphics[width=0.92\textwidth]{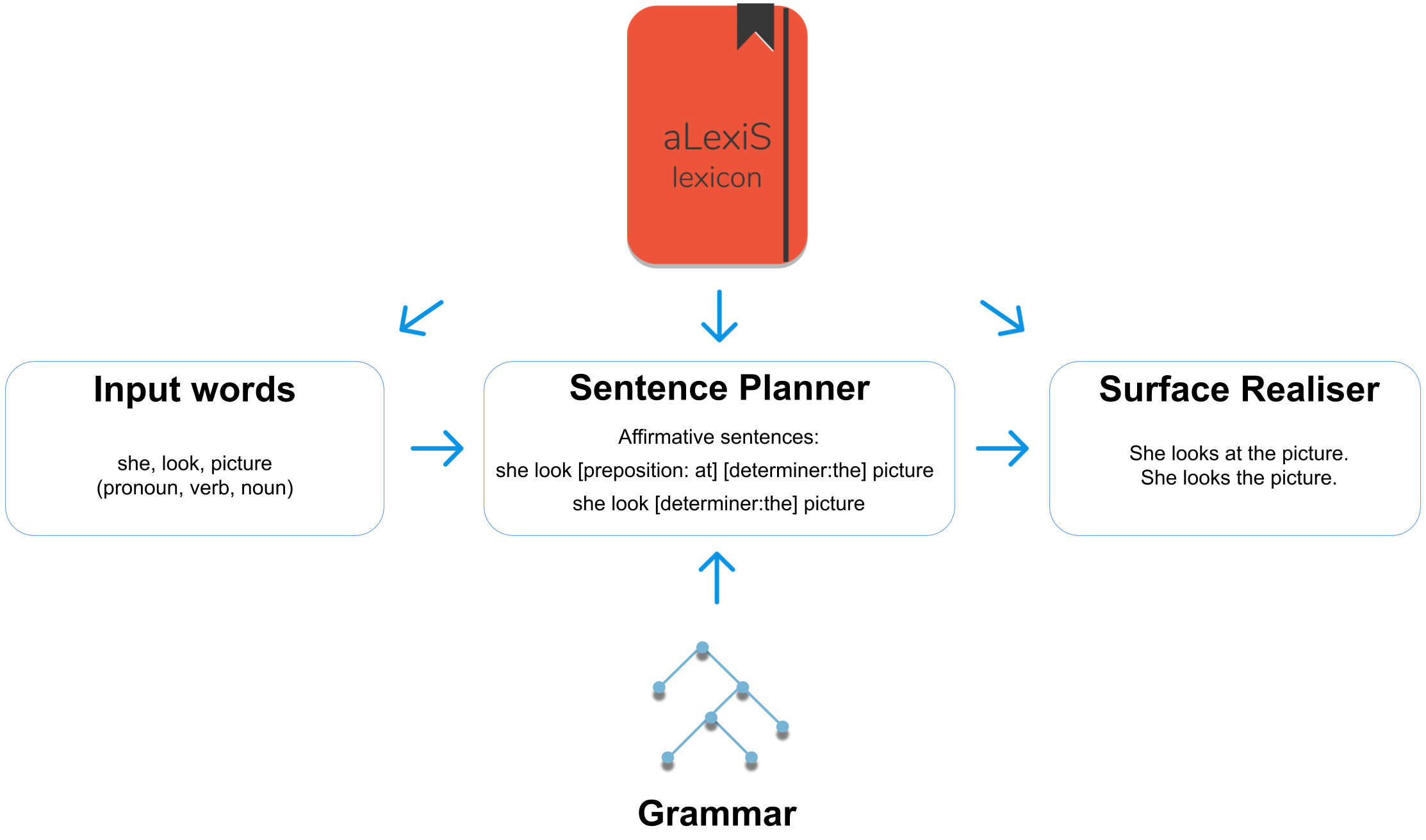}
\caption{Sentence generation with our two-stage NLG library}
\label{generation_diagram}
\end{figure*}

\subsubsection{Detection of the linguistic structure  (affirmative, negative or interrogative) of the sentence (Sentence Planner)}
The sentence is considered negative if one of the input words is the negation adverb `not'. It is treated as interrogative if the input words include a question mark. If both elements are present, a negated question is generated by the system. The sentence is considered affirmative in any other case. This is the case in our example in Fig. \ref{generation_diagram}. The library also adds extra elements corresponding to any linguistic realisations in the grammar and the knowledge in aLexiE.

\subsubsection{Subject insertion (Sentence Planner)}
Imperative sentences and other sentences with elided subjects are quite common in English, for example `Go to your room'. We want the {\sc nlg} process to be almost transparent to end users and, thus, if the user does not provide a subject, the library takes the personal pronoun `I' as such. Besides, it generates a second option with elided subject in case the user wants to create an imperative clause. In our example in Fig. \ref{generation_diagram} and Fig. \ref{flowchart}, the library does not include a subject because the user introduces `she' as input. This is detected thanks to the grammar.

\subsubsection{Inference of syntax structure (Sentence Planner)}
The separation between subject and predicate simplifies the identification of the best syntactic trees for the input words, since they are smaller. Once the Sentence Planner decides the type of sentence, the library sets the boundary between the subject and the predicate taking the main verb as a reference and then searches for the best syntactic structure that matches them. For this purpose we follow a Depth-First Search ({\sc dfs}) \cite{tarjan1971depth} in our grammar departing from the input words. In case that some of the input words are not in the aLexiE lexicon, they will be treated as proper nouns. This means no inflections will be applied to them.

In our example in Fig. \ref{syntaxTree}, the system infers two possible syntax structures (options 1 and 2).

\subsubsection{Inclusion of extra elements (Sentence Planner)}
Once the syntactic structure is decided, some extra elements such as determiners, prepositions, and conjunctions may be necessary. These elements are inserted in the sentence if they are associated to feasible grammar realisations. In our example in Figs. \ref{generation_diagram} and \ref{flowchart}, the library adds to the output the extra elements that were inferred in the previous stage from the grammar.

\subsubsection{Morphological inflections (Realiser)}
This encompasses the inflections that are necessary to produce a grammatically correct sentence, in which the subject dictates the morphological conjugation, person, number, and gender inflections of the verb and other components. 

The library distinguishes between subject and predicate before generating a sentence. In this regard, it can apply linguistic features to adjust person, gender, and number, to create sentences with coordinate subject. For example, given the input words `caregiver', `I', `eat', `apples', the subject of the resulting sentence `The caregiver and I eat apples' is compound.

First, person, gender, and number features must be derived from the input words. The subject (expected to be a nominal or coordinated nominal syntagm) determines them. Continuing with the running example `The caregiver and I eat apples', the subject is a coordinated nominal syntagm. The first nominal sytagm within is composed of the determiner `the' and the singular noun `caregiver'. The second is the pronoun `I'. Consequently, the person and number of the sentence are first person and plural, and the verb `eat' is inflected accordingly.

\begin{table*}[ht!]
\centering
\caption{\label{functionalitiesExample} Results of our NLG library}
\begin{tabular}{ll}
\hline
\bf Input words & \bf Best output sentence/s\\ \toprule
something, be, not, right & Something isn't right.\\\hline
where, my, glasses, be, ? & Where are my glasses?
\\\hline
dinner, be, good, last, night & Dinner was good last night.\\\hline
appreciate, your, help, concern & I appreciate your help and concern.\\\hline
live, yellow, house & I live in the yellow house.\\\hline
how much, stamps, be, these, days, ? & How much are stamps these days?\\\hline
final, grades, be, available, after, class, today, ? & Are final grades available after class today?\\\bottomrule
\end{tabular}
\end{table*}

\begin{table*}[!ht]
\centering
\caption{\label{table-results} aLexiE lemma and form extraction statistics}
{\begin{tabular}{{l}{c}{c}{c}{c}}\hline
\multicolumn{1}{c}{}& \multicolumn{2}{c}{{\bf Initial}}& \multicolumn{2}{c}{{\bf Extracted}}\\\cline{2-5}
& \bf Lemmas & \bf Forms & \bf Lemmas & \bf {\%Lemmas}\\ \toprule
\multicolumn{1}{l}{\bf EnLex} & 508,000 & 695,000 & 212,021 & 41.74\%\\
\multicolumn{1}{l}{{\bf NIH}} & 505,145 & 955,564 & 67,660 & 13.39\%\\
\multicolumn{1}{l}{\bf English-Freeling} & 37,000 & 68,000 & 14,368 & 38.83\%\\\bottomrule
\end{tabular}}
\end{table*}

By default, the first time the user introduces input words, the library takes masculine gender, singular number and first person. Then, using aLexiE, it adapts these features with grammar rules. For instance, if the subject is a coordinated nominal syntagm, the output sentence is plural. Regarding gender, the output sentence is only feminine if all subject components are so\footnote{Some gender-specific words, such as waiter and waitress, remain in modern English.}. The following rules are applied in strict order to adjust the person feature: (1) if the subject has an element that refers to the first person, the output sentence is adjusted to that person; (2) if the sentence has an element that refers to the second person with no relation with the first person, the output sentence is adjusted to second person; (3) finally, if the sentence has an element that refers to the third person with no relation with the first and second persons, the sentence remains in third person.

aLexiE contains the number and gender inflections of all lemmas and the person features of verbs and pronouns. Once these features are decided, they are applied to all word inputs. However, in case the subject is missing, default features must be set (as previously mentioned).

The verbal tense of the output sentence is present unless a time adverb or a time adverbial locution (e.g. last week) are provided. For example, for the adverb `yesterday', the tense of the output sentence is past. This linguistic information can also be found in aLexiE.

The library handles contraction spellings as well. We implemented those from a freely accessible list\footnote{Available at {\tt https://en.wikipedia.org/wiki/Wikipedia:
List\_of\_English\_contractions}, July 2019.}.

In case a word is missing in the lexicon, no related features are available or they cannot be inflected, our library treats the word as a proper noun. 

When generating a sentence, it is necessary to create its syntagms and join them while respecting their syntactic and semantic function. For example, to create `She looks at the picture' from the input words `she', `look' and `picture', it is necessary to generate the nominal syntagm the `picture' and integrate it into a prepositional syntagm as `at the picture'. It is also necessary to build the subject of the sentence `she' and a predicate with `look' as the main verb. Finally, it is necessary to integrate in the output sentence the subject and the predicate with the prepositional complement (`she looks at the picture'). All these stages are automatic, even the inclusion of the preposition, from the syntactic and semantic information in aLexiE.

Table \ref{functionalitiesExample} shows examples of automatically generated sentences with increasing linguistic complexity. Alongside each example we indicate the input words.

\section{Experimental results}
\label{evaluation}

First, we compared the aLexiE lexicon with other lexica from the state-of-the-art (Section \ref{result-built}). Then, we evaluated automatic English text expansion with our system (Section \ref{englishtextexpansion}), both directly (by regenerating corpus sentences) and manually (from annotations) using a widely used corpus in the {\sc nlg} field (Section \ref{aac-corpus}). To the best of our knowledge, there is no other system for automatic text expansion. Therefore, we have created the English version of a corpus (Section \ref{paralel-corpus}), to compare automatic text expansion in a multilingual scenario.

\subsection{Lexicon}
\label{result-built}

There are several resources for statistical natural language processing and corpus-based computational linguistics\footnote{E.g. {\tt https://nlp.stanford.edu/links/statnlp.html}, July 2019.}. The main differences between aLexiE and those freely available online resources are coverage, correctness and completeness of linguistic information (morphology, syntax, and semantics).

\begin{table*}[ht!]
\centering
\caption{\label{table-semantic} Semantic sources for entries tagged as nouns in aLexiE}
\begin{tabular}{ccccccc} \hline
\multirow{2}{*}{} & \multicolumn{2}{c}{\bf MCR} & \multicolumn{2}{c}{{\bf FrameNet}} & \multicolumn{2}{c}{{\bf Total}} \\ 
& {\bf Present} & {\bf Absent} & {\bf Present} & {\bf Absent} & {\bf Present} & {\bf Absent}\\ \toprule
\multicolumn{1}{l}{\bf Nouns} & 69,784 & 8,182 & 25 & 8,157 & 69,809 & 8,157\\ \bottomrule
\end{tabular}
\end{table*}

Table \ref{table-results} shows the information we combined from the selected resources to create aLexiE. According to \cite{sagot2018multilingual}, EnLex contains 508,000 unique lemmas, corresponding to 695,000 unique forms. We only extracted some entries, yielding 212,021 unique lemmas that correspond to 41.74\% of the extracted entries. {\sc nih} contains 505,145 unique lemmas and 955,564 inflected forms. In this case we only extracted 67,660 unique lemmas that correspond to 13.39\% of the lexicon. Specifically, we only extracted adjective, adverb, conjunction, determiner, preposition, and pronoun entries, since other entries had no associated morphological information. Freeling for English contains 37,000 unique lemmas, corresponding to 68,000 unique forms. We only extracted adjectives, adverbs, and verbs (for the same reason as for {\sc nih}), producing 14,368 unique lemmas corresponding to 38.83\% of the original set.

Table \ref{table-semantic} shows the sources for the semantic classification of the nouns in aLexiE. We only searched in FrameNet\footnote{Available at {\tt https://framenet.icsi.berkeley.edu/
fndrupal}, July 2019.} the lemmas of the nouns that were missing in {\sc mcr}. Table \ref{table-merged-verified} shows the amount of lemmas after merging and verification. Table \ref{table-results2} shows the lexical categories of the lemmas and forms in aLexiE. Most were tagged as nouns (77,966), yielding over 141,000 aLexiE inflected forms. Determiners and pronouns were revised manually to include plural forms for the lemmas `this' and `that'.

\begin{table}[!ht]
\centering
\caption{\label{table-merged-verified} Merged and verified lemmas in aLexiE by lexical category}
{\begin{tabular}{{l}{c}{c}}
\hline
\bf Category & \bf Merged & \bf Verified\\ \toprule
Noun & 81,380 & 77,966\\
Pronoun & 86 & 86\\
Verb & 11,602 & 10,903 \\
Adjective & 47,078 & 28,971\\
Adverb & 10,353 & 1,867\\
Conjunction & 99 & 70\\
Determiner & 98 & 98\\
Preposition & 140 & 140\\
\hline
 Total & 150,836 & 120,101 \\\bottomrule
\end{tabular}}
\end{table}

\begin{table}[!ht]
\centering
\caption{\label{table-results2} aLexiE lemmas and forms by lexical category}
{\begin{tabular}{{l}{c}{c}}
\hline
\bf Category & \bf Lemmas & \bf Forms\\ \toprule
Noun & 77,966 & 141,580\\
Pronoun & 86 & 92\\
Verb & 10,903 & 39,493 \\
Adjective & 28,971 & 28,971\\
Adverb & 1,867 & 1,867\\
Conjunction & 70 & 70\\
Determiner & 98 & 100\\
Preposition & 140 & 140\\
\hline
Total & 120,101 & 212,313 \\\bottomrule
\end{tabular}}
\end{table}

\subsection{English text expansion}
\label{englishtextexpansion}

As previously said, we evaluated our system by extracting keywords from sentences from a widely used corpus in the {\sc nlg} field (Section \ref{aac-corpus}) and an ad-hoc Spanish-English parallel corpus (Section \ref{paralel-corpus}). We evaluated output quality in terms of completeness, correctness and similarity to the original sentence.

We decided to discard some 
common state-of-the-art metrics such as {\sc rouge} \cite{lin2003automatic} and {\sc bleu} \cite{papineni2002bleu}, because they weakly reflect human assessment of {\sc nlg}, as discussed in \cite{novikova2017we}. 

\subsubsection{AAC corpus}
\label{aac-corpus}

Even though our approach may be used for general {\sc nlg} scenarios, we chose an {\sc aac} corpus\footnote{Available at {\tt http://aactext.org}, July 2019.} for our first evaluation due to the interest of {\sc aac} as a representative real application. Some {\sc aac} tools such as Talk Together\footnote{Available at {\tt https://acecentre.org.uk/resources/talk-
together}, July 2019.} and LetMe Talk\footnote{Available at {\tt http://www.utac.cat/descarregues/cace-
utac}, July 2019.} have small vocabulary packages with rigid interactions. None of them generates messages taking morphological, syntactic, and semantic information into consideration. The interest of {\sc nlg} for {\sc aac} is illustrated by several previous works \cite{Copestake96,moreda2015use,nikolopoulos2011robotic}. First we selected sentences without commas or hyphens, to ensure that there was a single sentence/idea in a clause (our system could handle multiple ideas as separate sentences). Since our system performs {\sc nlg} automatically, we selected sentences in present, past, and future tense because these can be inferred by time adverbial complements. We then filtered the result to obtain the main words (adjectives, adverbs, nouns, pronouns, proper nouns, and verbs). Next we lemmatised all those words but the nouns and pronouns. This was because if we lemmatised the latter, the system would have no clue to generate a sentence with a plural noun or pronoun since the features of these particular words are independent from other components of the sentence (conversely, adjectives depend on the noun they modify). For this purpose we used the Spacy syntactic parser. The resulting dataset had 1,869 English sentences and their main words.

\paragraph{Annotation}
\label{aac-results}

We introduced the main words of a target sentence into our automatic {\sc nlg} system and we studied the output sentences. In case of a full match between the target and generated sentences, automatic generation was considered totally successful. This happened for 1,315 sentences, 70.25\% of the total. The remaining 554 were manually inspected. 

Of these, 15 differed only in few capital letters. This was due to errors in the target and missing words in aLexiE that were treated as proper nouns. Our system correctly replaced words by proper nouns in eight sentences, such as `I need a new harry potter book', which was generated as `I need a new Harry Potter book'. Even though the matches were inexact, we consider these sentences success cases rather than failures. In five sentences the system failed to detect the lexical category of some input words due to missing data in aLexiE, as in `I'm itchy', which our system generated as `I'm Itchy' (words like itchy are neither present in aLexiE nor in {\sc mwd}). We did not consider these sentences failures because they were due to missing words in the dictionary. Finally, there were two sentences containing words without an aLexiE entry that indeed existed in {\sc mwd}: `Need a bigger size' was generated by our system as `Need a Bigger size', and `It is 2 o'clock' was generated as `It is 2 O'clock'. These were indeed failures of the system.

There were 22 sentences containing spelling mistakes in the target such as `I have an appoinment with the doctor today' (`appoinment' instead of `appointment'). 

Consequently, our system was able to generate 1350\footnote{1315 + 8 + 5 + 22 = 1350} correct sentences automatically, corresponding to 72.23\% of the total.

Finally, it was only necessary to evaluate 517 sentences manually. They were revised by five {\sc nlg} researchers from {\it atlanTTic}, University of Vigo, with English skills equivalent to C1 in the Common European Framework of Reference for Languages ({\sc cefr}) or 95 or above in the Test of English as a Foreign Language ({\sc toefl}). Table \ref{annotatedfeatures} shows the annotation options.

\begin{table}[ht!]
\centering
\caption{\label{annotatedfeatures} Annotation tags}
\begin{tabular}{ll}
\hline
\bf Feature & \bf Values \\ \toprule
\multirow{3}{*}{Error type} & morphological (a), syntactic (b),\\ 
& lexicon (c), grammar (d), target (e), \\
& lemmatiser (f) \\\hline
Evaluation & 0-5 \\\hline
Best generation & optional \\\hline
Generation suggestion & optional \\\bottomrule
\end{tabular}
\end{table}

Manual evaluation considered six error possibilities: morphological error (a), syntactic error (b), aLexiE error (c), grammar error (d), target error (e), and lemmatiser error (f) (Table \ref{annotatedfeatures}). The annotators also rated the quality of the generation from 0 (not generated) to 5 (full match between target and output)\footnote{0 and 5 ratings were automatically treated.}. Moreover, when the system presented different alternative outputs the annotator had to choose one. We noticed that, in some errors, except for {\sc svo} order the system would have succeeded in generating the targets. The annotators were requested to provide output suggestions in that situation.

The annotation task took two months. We handed instructions and examples to the annotators in advance to guarantee the consistency of the resulting corpus. The tests exploited various features of English grammar such as sentence type and constructions with different word categories. The annotation script returned an {\sc xml} file. Listing \ref{completeexample} shows an annotated sentence example.

The final results can be summarised as follows. Firstly, we must distinguish the cases when our {\sc nlg} system generated a single possibility from those with several output sentences. In the first case the error type was set by majority vote between the annotators. If the annotators did not agree, the sentence was tagged with {\it no consensus about error type}. The final rating of each output sentence was computed as the arithmetic average of annotator ratings. In the second case, first we checked if there was consensus in the {\it best realisation} field. Otherwise, the sentence was tagged with {\it no consensus about best realisation}. If the annotators provided suggestions of best realisations and there was a consensus about them, we tagged and rated the best output candidate as in the first case. Table \ref{scenarios} shows the distribution of the annotations.

\begingroup
\setlength{\xfigwd}{\textwidth}
\begin{lstlisting}[caption={\vspace{\hcaption}Annotation example}, label={completeexample}]
<?xml version="1.0" encoding="UTF-8"?>
<TAGGING>
 <CLAUSE>
  <TARGET>Dropped my change.</TARGET>
   <Generated_Clauses>
    <Clause>
     Drop my change.
     <Error>b</Error>
     <Rating>1</Rating>
    </Clause>
    <Clause>
     I drop my change.
     <Error>a</Error>
     <Rating>2</Rating>
    </Clause>
  </Generated_Clauses>
  <Best_realisation>2</Best_realisation>
  <Suggestion_for_Generation>
   I dropped my change yesterday.
  </Suggestion_for_Generation>
 </CLAUSE>
</TAGGING>
\end{lstlisting}
\endgroup

\begin{table}[ht!]
\centering
\small
\caption{\label{scenarios} Distribution of annotations of our dataset}
\begin{tabular}{lc}\hline
\bf Annotation result & \bf Sentence number \\ \toprule
\multicolumn{1}{l}{No consensus about best realisation} & \multicolumn{1}{c}{0} \\\hline
\multicolumn{1}{l}{No consensus about error type but } & \multirow{2}{*}{45} \\
\multicolumn{1}{l}{consensus about best realisation} & \\\hline
\multicolumn{1}{l}{Total consensus} & 472 \\\bottomrule
\end{tabular}
\end{table}

When the annotators agreed about error and best realisation, their average rating indicates that the information in the target could be understood from the generated sentence. This also happened when the annotators agreed about the best realisation but there was no consensus about error type. The annotators suggested 367 different alternative outputs, of which our library generated 160 correctly and automatically (43.597\%). The remaining 207 sentences (56.403\%) were not manually inspected. We suppose that many of these generated sentences might be considered appropriate as well.

The main mistakes were due to verbal tense adjustment, since many sentences were in past tense but did not have any time-related complement. Another common failure was adding a different preposition instead of the one in the target (in some cases, however, this change did not modify the meaning of the output sentence).

\paragraph{Evaluation agreement}

Manual evaluation was monitored with two recognized agreement metrics that yield robust estimations of the differences between annotators: {\it Alpha}-reliability \cite{krippendorff2012content, krippendorff2011computing} and accuracy.

When the annotators perfectly agree, $Alpha = 1$. When their agreement seems by chance $Alpha = 0$. Obviously both extremes should be avoided.

Our evaluation focused on nominal data because we measured error annotation agreement between five observers. As previously said, we computed the agreement in error type and obtained the average rate. The first step was to build a 5-observers-by-523-sentences reliability data matrix containing $ 5\times 523 $ values.

Table \ref{reliabilityMatrix} shows that our system generated 523 sentences for the 517 target sentences in the corpus. This was because there were several generated candidates for some targets.

\begin{table}[ht!]
\centering
\caption{\label{reliabilityMatrix}Reliability data matrix of the AAC annotated dataset considering the error types in Table \ref{annotatedfeatures}}
\begin{tabular}{lcccccc}
\hline
{\multirow{1}{*}{}} & \multicolumn{6}{c}{\bf Sentences} \\
\cline{2-7}
\multicolumn{1}{l}{} & \bf 1 & \bf 2 & \bf ... & \bf \bf 125 & \bf ... & \bf 523 \\ \toprule
\multicolumn{1}{c}{\bf Annotator 1} & b & b & ... & a & ... & a \\
\multicolumn{1}{c}{\bf Annotator 2} & b & b & ... & d & ... & a \\
\multicolumn{1}{c}{\bf Annotator 3} & b & d & ... & d & ... & a \\
\multicolumn{1}{c}{\bf Annotator 4} & b & b & ... & b & ... & a \\
\multicolumn{1}{c}{\bf Annotator 5} & b & b & ... & b & ... & a \\\bottomrule
\end{tabular}
\end{table}

Next we tabulated the coincidence matrix in Table \ref{coincidenceMatrix}, in units. Coincidence matrices take into account the values in a reliability data matrix. They differ from contingency matrices in that the latter consider units in two dimensions, not values. Our coincidence matrix accounted for all pairable errors from the five annotators into a 6-by-6 square matrix, omitting references to annotators. This type of matrix is symmetric with respect to its main diagonal, which holds all perfect matches. Note that the coincidences were counted twice in the coincidence matrix. Disagreements (represented by off-diagonal cells) were also counted twice, yet in different cells.

\begin{table}[!ht]
\centering
\caption{\label{coincidenceMatrix}Coincidence matrix of our annotated dataset considering the error types in Table \ref{annotatedfeatures}} 
\begin{tabular}{ccccccc}
\hline
\multicolumn{1}{l}{} & \bf a & \bf b & \bf c & \bf d & e & \bf f \\ \toprule
\bf a & 698 & 12.5 & 3 & 46.75 & 39.5 & 8.25 \\
\bf b & 12.5 & 378 & 7.5 & 91 & 17.75 & 7.25 \\
\bf c & 3 & 7.5 & 37 & 10.75 & 3.5 & 14.25 \\
\bf d & 46.75 & 91 & 10.75 & 635.5 & 83.25 & 39.75 \\
\bf e & 39.5 & 17.75 & 3.5 & 83.25 & 55.5 & 18.5 \\
\bf f & 8.25 & 7.25 & 14.25 & 39.75 & 18.5 & 4\\\bottomrule
\end{tabular}
\end{table}

We then estimated inter-agreement accuracy between pairs of annotators. This simply averages the proportions given by the diagonal of the coincidence matrix. Note that it neither accounts for fortuitous (dis)agreement nor for value ordering. The results for $Alpha$ and accuracy in Table \ref{alphaaccuracyresults} are promising. Tables \ref{alphaMatrix} and \ref{accurayMatrix} represent inter-agreement between pairs of annotators \cite{dorussen2005assessing,poesio2005reliability,pestian2012sentiment}.

\begin{table}[t!]
\centering
\caption{\label{alphaaccuracyresults} Overall inter-annotator agreement}
\begin{tabular}{lllll}
\hline
\bf Agreement measure & \bf Value & & & \\ \toprule
\bf $Alpha$ & 0.582 & & & \\
 Accuracy & {0.691} & & & \\\bottomrule
\end{tabular}
\end{table}

\begin{table*}[!ht]
\centering
\caption{\label{alphaMatrix}$Alpha$ between annotator pairs}
\begin{tabular}{lccccc}\hline
\multicolumn{1}{c}{} & \multicolumn{1}{c}{\bf Annotator 1} & \multicolumn{1}{c}{\bf Annotator 2} & \multicolumn{1}{c}{\bf Annotator 3} & \multicolumn{1}{c}{\bf Annotator 4} & \multicolumn{1}{c}{\bf Annotator 5} \\\toprule
\multicolumn{1}{c}{\bf Annotator 1} & \multicolumn{1}{c}{-} & \multicolumn{1}{c}{0.726} & \multicolumn{1}{c}{0.485} & \multicolumn{1}{c}{0.675} & \multicolumn{1}{c}{0.649} \\
 \multicolumn{1}{c}{\bf Annotator 2} & 0.726 & \multicolumn{1}{c}{-} & \multicolumn{1}{c}{0.451} & \multicolumn{1}{c}{0.614} & \multicolumn{1}{c}{0.646} \\
 \multicolumn{1}{c}{\bf Annotator 3} & 0.485 & 0.451 & \multicolumn{1}{c}{-} & \multicolumn{1}{c}{0.508} & \multicolumn{1}{c}{0.447} \\
 \multicolumn{1}{c}{\bf Annotator 4} & 0.675 & 0.614 & 0.508 & \multicolumn{1}{c}{-} & \multicolumn{1}{c}{0.624} \\
 \multicolumn{1}{c}{\bf Annotator 5} & 0.649 & 0.646 & 0.447 & \multicolumn{1}{c}{0.624} & \multicolumn{1}{c}{-} \\\bottomrule
\end{tabular}
\end{table*}

\begin{table*}[t!]
\centering
\caption{\label{accurayMatrix}Accuracy between annotator pairs}
\begin{tabular}{lccccc}\hline
\multicolumn{1}{c}{} & \multicolumn{1}{c}{\bf Annotator 1} & \multicolumn{1}{c}{\bf Annotator 2} & \multicolumn{1}{c}{\bf Annotator 3} & \multicolumn{1}{c}{\bf Annotator 4} & \multicolumn{1}{c}{\bf Annotator 5} \\\toprule
\multicolumn{1}{c}{\bf Annotator 1} & \multicolumn{1}{c}{-} & \multicolumn{1}{c}{0.803} & \multicolumn{1}{c}{0.608} & \multicolumn{1}{c}{0.759} & \multicolumn{1}{c}{0.751} \\
 \multicolumn{1}{c}{\bf Annotator 2} & 0.803 & \multicolumn{1}{c}{-} & \multicolumn{1}{c}{0.585} & \multicolumn{1}{c}{0.717} & \multicolumn{1}{c}{0.753} \\
 \multicolumn{1}{c}{\bf Annotator 3} & 0.608 & 0.585 & \multicolumn{1}{c}{-} & \multicolumn{1}{c}{0.621} & \multicolumn{1}{c}{0.587} \\
 \multicolumn{1}{c}{\bf Annotator 4} & 0.759 & 0.717 & 0.621 & \multicolumn{1}{c}{-} & \multicolumn{1}{c}{0.728} \\
 \multicolumn{1}{c}{\bf Annotator 5} & 0.751 & 0.753 & 0.587 & \multicolumn{1}{c}{0.728} & \multicolumn{1}{c}{-} \\\bottomrule
\end{tabular}
\end{table*}

\subsubsection{Spanish-English parallel corpus}
\label{paralel-corpus}

We are not aware of the existence of other systems for automatic text expansion as we have defined it. Therefore, we decided to apply our automatic {\sc nlg} system to Spanish and English and compare the results.

For this purpose, we manually created the English version of the Spanish corpus used in \cite{garcia2019library}\footnote{Available at {\tt https://www.gti.uvigo.es/images/manual\_
evaluation\_EN.txt}, July 2019.}. 

The final parallel corpus is composed of 948 English/Spanish sentences covering various grammar features such as different sentence types and constructions with different word categories.

Table \ref{comparison-nlg-simplenlg} shows a comparison between the English and Spanish versions in terms of automatic generation using the parallel corpus. First, our system generated 613 English sentences automatically. The remaining 123 sentences were inspected manually. We noticed that the most relevant mistakes (in 106 sentences) were due to Wikipedia training, since the system failed to add a certain preposition in the target. Two of the other 17 sentences were actually correct, since they only differed in some capital letters, and four had differences in determiners that did not affect their meaning. 

Second, our system was able to automatically generate 72 English sentences out of the 212 that the system in \cite{garcia2019library} failed to generate in Spanish. The remaining 140 sentences were manually inspected. The most common mistakes were related to verbal tense adjustment (64 sentences) and wrong prepositions (34 sentences). 

Next we compared the approach in this paper with the system in \cite{garcia2019library}. We correctly generated 77.64\% and 72.26\% of the Spanish and English sentences in the parallel corpus, respectively.

The most important difference between the two languages was the use of prepositions. In Spanish there were few mistakes of this kind, but they were the most common in English. In our opinion this was due to the difficulty to detect phrasal verbs.

\begin{table}[ht!]
\centering
\caption{\label{comparison-nlg-simplenlg}Comparison between the Spanish ({\sc ES}) and English ({\sc EN}) systems using the parallel corpus (automatic evaluation)} 
\begin{tabular}{lcccc}\cline{2-5}
\multicolumn{1}{c}{\multirow{1}{*}{}} & \multicolumn{1}{l}{} & \multicolumn{3}{c}{\bf ES version} \\
\multicolumn{1}{c}{\multirow{1}{*}{}} & \multicolumn{1}{l}{} & \bf Automatic & \multicolumn{1}{c}{\bf Manual} & \multicolumn{1}{c}{\bf Total} \\\toprule
\multirow{3}{*}{\bf EN version} & \multicolumn{1}{c}{\bf Automatic} & 613 & \multicolumn{1}{c|}{72} & 685 \\
 & \multicolumn{1}{c}{\bf Manual} & 123 & \multicolumn{1}{c|}{140} & 263 \\ \cline{2-5} 
 & \multicolumn{1}{c}{\bf Total} & 736 & \multicolumn{1}{c|}{212} & 948\\\bottomrule
\end{tabular}
\end{table}

\section{Conclusions}
\label{conclusions}
We have developed an automatic hybrid system for English text expansion. Relying on the aLexiE lexicon and our grammar, our system is able to perform fully automatic text expansion from few input words. The integration of new lexical resources for any language is simple. The architecture separates domain-independent from domain-dependent components, so that the latter can be substituted. We remark that the aLexiE lexicon and the grammar we have developed for English expansion are relevant results in themselves. They could be useful to other {\sc nlg} researchers.

As far as we know this is the first fully automatic hybrid system for English text expansion, combining a knowledge base of vocabulary and grammar realisations with a statistical language model for preposition inference. Our system has a good success rate when generating coherent and grammatically correct sentences from user-selected input words.

The surface realiser relies on aLexiE and our grammar to take its decisions. For this reason, we provide a detailed description of the procedure to create them. As future work we plan to predict the best grammar realisation for input words without {\sc svo} order. 

We have offered an insightful analysis of semantic similarities and differences between target and output texts, and we have assessed system performance using state-of-the-art metrics. First, we evaluated the system by regenerating texts from an {\sc aac} corpus. The system succeeded 72.23\% of the time. As future work we will evaluate the system with {\sc aac} users taking advantage of the broad community of our Pictodroid suite\footnote{Available at {\tt http://www.accegal.org/en/pictodroid}, July 2019.}. Second, we conducted automatic text expansion in English and Spanish and compared the results. Due to the lack of a multilingual dataset, we had to create a parallel corpus for this purpose. Even though generation was slightly better for Spanish, thanks to its more predictable preposition usage, performances were comparable. 

As another future research line we plan to create similar systems for new languages such as French and Portuguese.

\bibliography{mybibfile.bib}{}
\bibliographystyle{IEEEtran}

\EOD

\end{document}